\documentclass[10pt,letterpaper]{article}
\usepackage[top=0.85in,left=2.75in,footskip=0.75in,marginparwidth=2in]{geometry}

\usepackage[utf8]{inputenc}

\usepackage{cite}
\usepackage{bm}

\usepackage{nameref,hyperref}

\usepackage[right]{lineno}

\usepackage{microtype}
\DisableLigatures[f]{encoding = *, family = * }

\raggedright
\usepackage{changepage}

\usepackage[aboveskip=1pt,labelfont=bf,labelsep=period,singlelinecheck=off]{caption}

\makeatletter
\renewcommand{\@biblabel}[1]{\quad#1.}
\makeatother

\usepackage{lastpage,fancyhdr,graphicx}
\usepackage{epstopdf}
\fancyheadoffset[L]{2.25in}
\fancyfootoffset[L]{2.25in}

\usepackage{color}

\definecolor{Gray}{gray}{.25}

\usepackage{graphicx}
\usepackage{multirow}

\usepackage{sidecap}

\usepackage{wrapfig}
\usepackage[pscoord]{eso-pic}
\usepackage[fulladjust]{marginnote}
\reversemarginpar

\begin{document}
\vspace*{0.35in}

\begin{flushleft}
{\Large
\textbf\newline{Interpretable Data Mining of Follicular Thyroid Cancer Ultrasound Features Using Enhanced Association Rules}
}
\newline
\\
Songlin Zhou\textsuperscript{1},
Tao Zhou\textsuperscript{2},
Xin Li\textsuperscript{3},
Stephen Shing-Toung Yau\textsuperscript{4,*}
\\
\bigskip
\bf{1} Qiuzhen Colledge, Tsinghua University, No.1 Qinghuayuan Street, Haidian District, Beijing, 100084, China
\\
\bf{2} Department of Mathematics, Tsinghua University, No.1 Qinghuayuan Street, Haidian District, Beijing, 100084, China
\\
\bf{3} Department of General Surgery, Peking University Third Hospital, No.49 North Garden Road, Haidian District, Beijing, 100191, China
\\
\bf{4} Beijing Institute of Mathematical Sciences and Applications(BIMSA), No.544 Hefangkou village Huaibei town, Huairou District, Beijing, China
\\
\bigskip
* yau@uic.edu

\end{flushleft}

\section*{Abstract}
\subsection*{Purpose}
Thyroid cancer has been a common cancer. Papillary thyroid cancer and follicular thyroid cancer are the two most common types of thyroid cancer. Follicular thyroid cancer lacks distinctive ultrasound signs and is more difficult to diagnose preoperatively than the more prevalent papillary thyroid cancer, and the clinical studies associated with it are less well established. We aimed to analyze the clinical data of follicular thyroid cancer based on a novel data mining tool to identify some clinical indications that may help in preoperative diagnosis.
\subsection*{Methods}
We performed a retrospective analysis based on case data collected by the Department of General Surgery of Peking University Third Hospital between 2010 and 2023. Unlike traditional statistical methods, we improved the association rule mining, a classical data mining method, and proposed new analytical metrics reflecting the malignant association between clinical indications and cancer with the help of the idea of SHAP method in interpretable machine learning.
\subsection*{Results}
The dataset was preprocessed to contain 1673 cases (in terms of nodes rather than patients), of which 1414 were benign and 259 were malignant nodes. Our analysis pointed out that in addition to some common indicators (e.g., irregular or lobulated nodal margins, uneven thickness halo, hypoechogenicity), there were also some indicators with strong malignant associations, such as nodule-in-nodule pattern, trabecular pattern, and low TSH scores. In addition, our results suggest that the combination of Hashimoto's thyroiditis may also have a strong malignant association.
\subsection*{Conclusion}
In the preoperative diagnosis of nodules suspected of follicular thyroid cancer, multiple clinical indications should be considered for a more accurate diagnosis. The diverse malignant associations identified in our study may serve as a reference for clinicians in related fields.

\section*{Introduction}

Thyroid carcinoma has become one of the most prevalent cancers worldwide \cite{pizzato2022epidemiological}. Thyroid carcinoma can be classified into three categories: differentiated, medullary and anaplastic \cite{chen2023thyroid}. Differentiated thyroid carcinoma is the most common \cite{fagin2016biologic} and has two main subtypes: papillary thyroid carcinoma (PTC) and follicular thyroid carcinoma (FTC). PTC can be considered the most widespread thyroid carcinoma, accounting for almost $80\%$ of all thyroid carcinoma cases \cite{chen2023thyroid}. FTC is the second most common thyroid cancer, accounting for about $10\%$ of the total number of cases \cite{zhang2024role}. FTC is more aggressive than PTC and thus has a worse prognosis. Unlike PTC, FTC does not have typical discriminatory signs and is difficult to diagnose preoperatively by imaging or case-puncture examination \cite{miyauchi2012prognostic}. The difficulty this poses is that for a follicular thyroid neoplasm, it is often difficult for the clinician to differentiate between malignant FTC and benign follicular thyroid adenoma (FTA) \cite{zhang2024role}: overestimating the risk of malignancy can lead to surgical over-excision of the thyroid gland, whereas underestimating the risk of malignancy can lead to surgical under-clearance of the lesion. Therefore, the preoperative diagnosis of FTC presents significant challenges.

There has been some clinical work examining the effect of different indications on follicular thyroid cancer, using mainly statistical methods (e.g., single- or multi-factorial logistic regression, LASSO regression \cite{yu2019development,tang2021development,li2023us}, etc.) and machine-learning methods (e.g., random forests, xgboost \cite{lin2024generating}, etc.). In terms of interpretability, which is of great concern in clinical practice, statistical methods mainly use some indicators such as odds ratio (OR) to explain the significance of variables, while machine learning methods mainly rely on some mainstream interpretability methods such as SHAP \cite{lundberg2017unified}.

Inspired by the principles of the SHAP method, we found that we could combine the idea of SHAP with a classic data mining method, association rule mining, to obtain more directly interpretable analyses of clinical data. The difficulties in preoperative diagnosis of follicular thyroid cancer itself have also aroused our interest in using new methods in practice.

\paragraph{Association Rule Mining}
Association rule mining is a classical data mining technique, proposed by Agrawal et al. in 1993 \cite{agrawal1993mining}. The technique was initially proposed to mine a database of customer transactions (each transaction consists of all the items purchased by a customer) to find potential correlations between the sales of different items, in order to better guide shopping malls in designing their marketing strategies. We first introduce several concepts in association rule mining:

\begin{itemize}
    \item Item: A clinical indication is expressed as an item, denoted by $i$. For example, “composition: solid” can be considered as a term.
    \item Itemset: A set consisting of a portion of items, denoted by $I$. A case may have more than one clinical indication, and these clinical indications constitute an item set. A set of k items is called a k-item set. For example, $\{$(Composition: solid), (Distribution of microflows: mainly central vascularity), (Nodule-in-nodule: yes)$\}$ is a 3-item set.
    \item Transaction: A record of clinical indications for a case stored in a hospital database is called a transaction, denoted by $t$. For example, a record of the last preoperative ultrasound, thyroid function, and puncture pathology performed on a case forms a transaction.
    \item Transaction set: A collection consisting of transactions, denoted by $\mathcal{T}$. For example, hospital-stored case data used in a retrospective study constitute a transaction set.
    \item Support of an itemset $I$: Represents the frequency of occurrence of item set $I$ in all transactions in the transaction set. For example, if the transaction set is 1000 case records, the itemset $I$ is $\{$(Composition: solid)$\}$, and 600 out of these 1000 case records are shown to have a solid composition, then the support of $I$ is $600/1000*100\% = 60\%$. The itemset with support higher than a human-specified threshold is called the frequent itemset.
    \item Association rule: An implication form like $X\to Y$ where $X,Y$ are two itemsets without intersection. The itemset to the left of the arrow is called antecedent, and the itemset to the right of the arrow is called consequent. Note that the correlation rule only indicates that there is a correlation between antecedent and consequent, and does not imply that there is a necessary causal relationship between them.
    \item Confidence: A rule $X\to Y$ has confidence $c$ in the transaction set $\mathcal{T}$ iff at least a $c$ proportion of all transactions in $\mathcal{T}$ that contain the itemset $X$ also contain the itemset $Y$. Confidence can be computed by $Conf(X\to Y)=Supp(X\cup Y)/Supp(X)$.
    \item Lift: lift is defined as $Lift(X\to Y)=Conf(X\to Y)/Supp(Y)$, which denotes the correlation between occurrences of itemset A and occurrences of itemset B. When its value is greater than 1, it means that the occurrences of itemset A and the occurrences of itemset B are positively correlated, when its value is less than 1 it indicates a negative correlation, and when its value is 1 it means that the two itemsets are independent.
\end{itemize}
A general association rule mining algorithm first identifies all the itemsets in the dataset whose support is not lower than a specified threshold $\eta_{supp}$ and whose number of items is greater than or equal to 2, and then constructs an association rule based on these itemsets whose confidence level is not lower than a specified threshold $\gamma_{conf}$.

However, there are many issues to overcome to run association rule mining on the dataset used in this study:
\begin{enumerate}
    \item Data conversion. Since there are some continuous variables (e.g., mean diagram of nodules, nodule volume, metabolic indexes, etc.) in the clinical data records, we need to transform them into categorical variables in order to construct a dataset suitable for association rule mining. In many studies, this transformation is realized based on clinical knowledge, but not all continuous-type variables have this clinical consensus.
    \item Threshold selection. There are opposing considerations for the choice of both $\eta_{supp}$ and $\gamma_{conf}$: (1) The support threshold $\eta_{supp}$ controls how “common” the mined frequent itemsets are in the transaction set. If the support threshold is too low, the time complexity of mining frequent itemsets will be higher on the one hand, and on the other hand, the mined frequent itemsets may be too rare, which may affect the validity of subsequent association rules based on frequent itemsets. However, if $\eta_{supp}$ is chosen too high, many potentially research-useful itemsets may be overlooked, and thus some interesting results may be lost in the subsequent construction of association rules (for example, there may be clinical indications in the dataset that are otherwise uncommon that happen to be of interest to us), which is undesirable in retrospective studies where exploring the data and generating hypotheses are the main goals. (2) The confidence threshold $\gamma_{conf}$ controls the reliability of the constructed association rules. If the threshold is chosen too low, although a larger number of rules can be generated so as not to miss some potentially interesting associations, a large number of low-confidence rules can be generated, causing interference in the subsequent analysis; while if $\gamma_{conf}$ is chosen too high, although the reliability of the association rules is more adequately safeguarded, at the same time it may also dramatically reduce the number of rules generated, or even only get some already consensus conclusion.
    \item Filtering of itemsets and rules. Classical association rule mining algorithms do not impose additional restrictions on the itemsets and rules, and if directly applied to clinical data analysis, a large number of meaningless rules will be generated, which brings a lot of redundant computations, as well as additional trouble for the subsequent analysis of association rules. In our study, since we are only concerned with the association between clinical indications and nodal malignancy, we add filtering rules to our algorithm during frequent itemset mining and association rule generation, i.e., we only consider frequent itemsets containing “malignant” and only generate association rules with “malignant” as the consequent. This improves the efficiency of the algorithm and reduces the difficulty of analyzing the results.
    \item Analysis of the rules. If the dataset is small (few case records, or few clinical indications considered), then the mined rules can be analyzed directly on a rule-by-rule basis, from which research-useful rules can be selected. However, in our study, a total of 2091 case records and 23 kinds of clinical indications (corresponding to 102 items) are involved, so that the number of generated rules is so large that it is impossible to analyze them one by one, which requires us to devise some methods and metrics to get valuable information from them.
\end{enumerate}

We try to provide some solution ideas to several difficulties in the practical application of association rule mining mentioned above and apply them to mining clinical indicators with high follicular thyroid cancer associations, with a view to enriching the methodology of clinical data analysis while providing reference to fellow researchers of follicular thyroid cancer. We try to provide some solution ideas to several difficulties in the practical application of association rule mining mentioned above and apply them to mining clinical indicators with high follicular thyroid cancer associations, with a view to enriching the methodology of clinical data analysis while providing reference to fellow researchers of follicular thyroid cancer. Unlike some previous work on clinical data analysis based on association rule mining \cite{mohamed2023knowledge,sankar2023association}, we do not analyze the high-confidence rules themselves, but rather draw on the rules to reflect the role of the clinical indications themselves. Our proposed method does not rely on model assumptions, is highly interpretable, and is able to intuitively demonstrate potential patterns in the data, and is therefore well suited for use in retrospective studies as a supplement to commonly used methodologies.

\section*{Materials and Methods}

\subsection*{Data Preprocessing}
To analyze the dataset using association rule mining based on apriori algorithm, we need to transform the dataset into a suitable format, specifically, each clinical indication needs to be transformed into discrete forms of categorical variables. This poses a problem in that we need to bin the data for continuous variables. The variables to be binned included max diagram, mean diagram, BMI, age, mean TSH score (the time-interval adjusted logTSH, see \cite{ito2023thyroid}), TSH tRMSSD (the time-adjusted root mean square of successive differences of logTSH, see \cite{taquet2023early}). Combining clinical experience and general consensus, we adopted a mixed supervised and unsupervised data binning method.
\begin{itemize}
    \item Supervised binning: Max diagram, mean diagram, BMI and age were included. For max  diagram and mean diagram, taking into account clinical experience, especially the ease of use in clinical judgment, we directly categorize the data into five categories at 1 cm intervals: less than 1 cm, 1-2 cm, 2-3 cm, 3-4 cm and not less than 4 cm. BMI was categorized into four categories according to the industry standard of the People's Republic of China (WS/T 428-2013): BMI $<$ 18.5 was considered underweight, 18.5 $\leq$ BMI $<$ 24 was considered normal weight, 24 $\leq$ BMI $<$ 28 was considered overweight, and BMI $\ge$ 28 was considered obese. For age, we divided the boxes into 10-year intervals based on the age spread of the dataset.
    \item Unsupervised binning: for mean TSH score and TSH tRMSSD, we use unsupervised binning based on kmeans \cite{lloyd1982least,macqueen1967multivariate}. Specifically, after normalizing the data with non-N/A values between -1 and 1, the kmeans algorithm is run to divide the data into 5 clusters. The selection of k=5 is based on the following considerations: if the value of k is too large, the number of categories of the binned variables is too large, which will slow down the efficiency of association rule mining; if the value of k is too small, the data binning may not work well. In addition, for mean TSH score, we did not observe a significant distributional bias, so we performed the binning algorithm directly on the original values, but for TSH tRMSSD, we observed a significant right-skewed distribution, and considering the existence of values very close to zero for this variable, we first performed a logarithmic transformation with an offset:
    \begin{equation}
        y=\log(x+0.00001)
    \end{equation}
\end{itemize}

In addition, for each feature, where the N/A value of the data (if present) is treated as a separate category. The vast majority of features were not data-populated, with the exception of three indicators known to have a clear correlation with malignancy (envelope invasion, vascular invasion, and tumor envelope completeness), because we wanted to mine the raw data as faithfully as possible to reflect the situation rather than adding too many artificial assumptions to create a model speculation. No additional binning was done for the remaining clinical indications that were themselves categorical variables. After data binning, including the degree of benignity or malignancy of the nodule, we obtained a total of 102 data bins (including 19 missing value bins), which corresponded to 22 clinical indicators and 1 categorical label (benign and malignant).

We then sequentially numbered all the features to build a data structure that conforms to the apriori class algorithm. We used a custom apriori family of functions rather than python's library functions. Specifically, we added filtering rules to the apriori algorithm's frequent item set generation process:
\begin{enumerate}
    \item make sure that the prior does not contain any of the 19 N/A bins;
    \item ensure that the posterior itemset contains only the term ‘malignant’ and no other items;
\end{enumerate}
this reduces the uncertainty of the mined rules and improves the efficiency of the algorithm.

The conclusions of the article are based on $\eta_{supp} = 10/1673$, under which parameter choice we mined 518,867 frequent itemsets (specifically, these contain 2-item sets to 12-item sets, numbering 67, 1211, 8287, 27467, 49713, 52443, 32883, 11816, 2228, 172, 2), and then based on those frequent itemsets an equal number of rules were generated.

\subsection*{Improvements for association rule mining}
In this subsection, we provide methodological level answers to several questions raised in the introduction section about association rule mining.
\subsubsection*{Threshold Selection}
As we mentioned earlier, the support and confidence based association rule mining algorithm relies on a minimum support threshold $\eta_{supp}$ and a minimum confidence threshold $\gamma_{conf}$. We give the following considerations regarding the selection of these two thresholds: for support, if $\eta_{supp}$ is too low, we may generate more rules, but the validity of the rules will be reduced because they may come from incidental or coincidental cases. Here we specify multiple hard thresholds: 5/1673, 10/1673, which means “at least 5 or 10 case records have the combination of clinical indications implied by the item set”, respectively. And regarding confidence, a lot of work has been devoted to finding those rules with “high confidence”, but since association rule mining itself cannot infer causality, it is difficult to judge how “high” the confidence should be. In this study, we are not committed to finding high-confidence rule itself, but rather we wish to measure the association between clinical indications and nodal malignancy with the help of rules. Briefly, we consider the “confidence change ratio” of a specific indication for nodal malignancy, where a specific indication can improve the confidence to at most 1, based on the support for the single item set of malignancy in the dataset $\beta=259/1673$, which corresponds to confidence change ratio $1/\beta$. Correspondingly, then, for the case of reduced confidence, we consider at most that the confidence is reduced by the same ratio $1/\beta$, i.e., the minimum confidence is $\gamma_{conf}=\beta/\left(1/\beta\right)=\beta^2$.
\subsubsection*{Algorithm Optimization}
We use apriori algorithm for association rule mining. Although the size of the dataset is not large, 23 indicator types and 102 indicator categories (bins in data binning, or items in association rule mining) may generate a very large number of itemsets, then the traditional apriori algorithm will be severely limited by the efficiency bottleneck. For this reason, we have optimized the algorithm in various ways. Specifically:
\begin{enumerate}
    \item Pruning based on items of interest: Since we only focus on the correlation between clinical indications and nodal malignancy, when we use apriori algorithm to iteratively construct frequent itemsets, we can keep only those itemsets that contain “malignancy” from the first step of searching for frequent binomials. In addition, in the rule generation process, we only consider rules with a single “vicious” item as the posterior item set, so that a frequent item set will only generate one rule at most, which greatly improves the efficiency of rule generation, and also realizes the effect of filtering rules. In addition, during the frequent itemset generation process, we filter out the itemsets that contain “missing boxes”, which reduces the bias for further analysis of the subsequently generated rules.
    \item The use of prefix dictionaries: In the iterative generation of k-frequent itemsets from k-1-frequent itemsets, a prefix dictionary is used as a data structure to organize the candidate itemsets with the same first k-2 elements together, so that only these candidate itemsets with the same prefixes need to be combined collectively and judged whether they form a k-frequent itemset or not.
    \item Complexity tradeoffs in support computation: The apriori algorithm involves the calculation of support for a large number of candidate itemsets, here we have tested and adopted a way to optimize the time complexity at the expense of space complexity, using a hash table to store the support that has been calculated in order to improve the efficiency of the program.
\end{enumerate}
\subsubsection*{Analysis of Rules}
Analyzing millions of rules one by one is complex, but we can use the rules as a tool to inversely analyze the correlation between clinical indications and nodal malignancy. With the help of the concept of lift in association rule mining, we define the following metrics to measure the strength of association between a particular clinical indication and nodal malignancy: Let the mined rules constitute a set $\mathcal{R}$. For a particular clinical indicator $i$, $(r_1,r_2)$ is said to be a “rule-of-interest pair” if the symmetric difference of the set of antecedents $A(r_1),A(r_2)$ of the two rules $r_1,r_2\in\mathcal{R}$ is $\{i\}$. In other words, we look for pairs of rules among the mined rules that have only one difference between the presence or absence of item $i$ in the set of antecedents. Denote the full set of interest rule pairs for $i$ as the set $\mathcal{RP}_i$, then we define 
\begin{itemize}
    \item the confidence ratio (CR) of $i$ for nodal malignancy as
        \begin{equation}
            CR(i)=Conf(r_{w/(i)}^j)/Conf(r_{w/o(i)}^j),\quad (r_{w/(i)}^j,r_{w/o(i)}^j)\in\mathcal{RP}_i,
        \end{equation}
        where $r_{w/(i)}^j$ denotes the rule whose antecedent set contains $i$ in the interest rule pair, and $r_{w/o(i)}^j$ denotes the rule whose antecedent set does not contain $i$.
    \item the average confidence boost (ACB) of $i$ for nodal malignancy as
        \begin{equation}
            ACB(i)=\exp\left\{ \frac{1}{|\mathcal{RP}_i|} \sum_{(r_{w/(i)}^j,r_{w/o(i)}^j)\in\mathcal{RP}_i}\big[\log[Conf(r_{w/(i)}^j)]-\log[Conf(r_{w/o(i)}^j)]\big]\right\},
        \end{equation}
         Here $ACB(i)$ computes, in effect, the geometric mean of the ratio of the confidence of the rules whose antecedent set contains $i$ to the confidence of the rules whose antecedent set does not contain $i$ over all pairs of rules in the $\mathcal{RP}_i$.
    \item the proportion of rules with increased confidence (PIC) with $i$ as
        \begin{equation}
            PIC(i)=\frac{1}{|\mathcal{RP}_i|} \sum_{(r_{w/(i)}^j,r_{w/o(i)}^j)\in\mathcal{RP}_i}\bm{1}\{r_{w/(i)}^j>r_{w/o(i)}^j\}.
        \end{equation}
        Here the PIC reflects in what proportion of rule-of-interest pairs the clinical indication $i$ acts as a confidence enhancer.
\end{itemize}
These two metrics allow us to screen for clinical indications with a clear malignant association and to test the validity of the methodology by comparing it with existing clinical findings.

\section*{Results}

\subsection*{Study Population}
We conducted a retrospective study based on data from 3573 cases recorded between March 2010 and November 2023 at the Department of Thyroid Surgery, Peking University Third Hospital. Data records are retrieved by ‘patient id - nodule id’, so here a nodule is called a case. A data record contains ultrasound, pathology, and thyroid function tests. There may be multiple records for the same case, and in this study, we kept only the records closest to the time of surgery for this case, to ensure that these tests were closest to the real situation corresponding to the benign and malignant labels of the postoperative pathological feedback. We did not conduct a cohort study in the time dimension; for most clinical indications, we used only the record of the closest examination to the time of surgery to reduce possible association bias. Only two indicators related to thyroid-stimulating hormone (TSH) were calculated using time series:
\begin{itemize}
    \item Mean TSH score: an indicator of the average level over time. This indicator is calculated as follows:
        \begin{enumerate}
            \item LogTSH is calculated by filtering out records with TSH levels of N/A, taking into account the records of previous examinations of the case recorded in the hospital;
            \item For each nodule, the average of the two adjacent logTSHs is calculated, and these averages are weighted in the time dimension according to the time interval;
            \item For cases with only one examination record, the logTSH of the only inspection record is retained.
        \end{enumerate}
    \item TSH tRMSSD: An indicator of the level of volatility over time. This indicator is calculated as follows:
        \begin{enumerate}
            \item LogTSH is calculated by filtering out records with TSH levels of N/A, taking into account the records of previous examinations of the case recorded in the hospital;
            \item For each node, calculate the slope of the logTSH at two adjacent time points (divide the change in logTSH by the time interval);
            \item Calculate the mean of the squares of the slopes over all consecutive time intervals and then take the square root.
        \end{enumerate}
\end{itemize}
Cases in the dataset were classified by postoperative pathology into three types: benign, malignant potential undetermined (MPU), and malignant. We did not take into account the MPU cases in our analyses, mainly due to the fact that our method is a data-mining one, and the inclusion of such data with ambiguous conclusions in the analytical framework could have a negative impact on the results. In addition, there was one record in the data where the tumor envelope condition indicator was outside the limit of the indicator, and we were unable to determine whether this stemmed from a data entry error and therefore chose not to include this record in the analysis. After the above screening process, we retained 1673 case records.

\subsection*{Identification of high malignant correlation indicators}
As a practice of the above methodology, we analyzed the data set on clinical indications of follicular thyroid cancer. Association rule mining is traditionally used to find rules of interest, in the scenario of this study, i.e., to find combinations of clinical indications that have a high risk of malignancy. However, as we mentioned in the introduction section, there can be so many such rules that it is not so convenient to analyze the rules themselves directly, so we start with the indications themselves and use all the mined rules as a bridge to look at the malignancy risk of the different indications.
From the results of Figure 1 and Table 1, it is clear that the ‘invasion indicators’ that are directly related to the clinical diagnosis of follicular thyroid cancer indeed have the highest ACBs, while the other signs that are more interesting to us include
\begin{enumerate}
    \item margin: lobulated, irregular;
    \item echogenic foci: peripheral calcifications, macrocalcifications, microcalcifications;
    \item present trabecular pattern or nodule-in-nodule pattern;
    \item relatively low mean TSH score;
    \item mean diagram or max diagram is no less than 4cm;
    \item halo: uneven thickness halo;
    \item echogenicity: hypoechoic;
    \item BMI: overweight;
    \item present Hashimoto's thyroiditis;
    \item vascularity: mainly central vascularity;
    \item high TSH tRMSSD
\end{enumerate}

\begin{figure}[!htbp] 
\includegraphics[width=\textwidth]{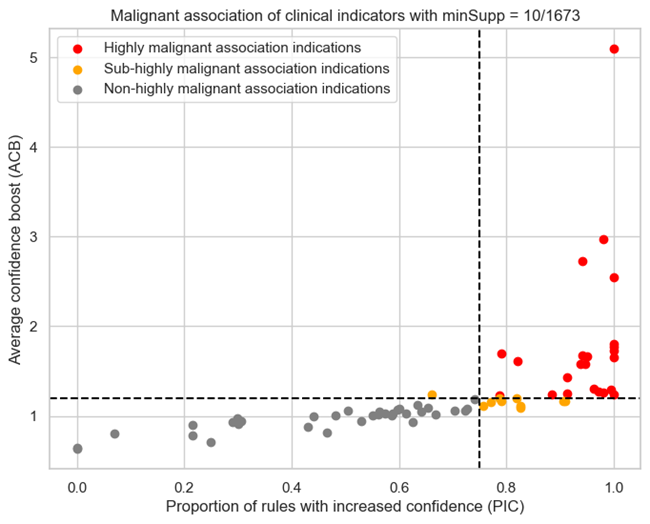}
\caption{\color{Gray} \textbf{Identification of malignant association of clinical indicators} This figure plots the scatterplot of different clinical indications using PIC and ACB (both PIC and ACB are defined in the “Analysis of Rules” subsection) as the horizontal and vertical axes and delineates the malignancy risk hierarchy for different indications by setting thresholds for PIC and ACB (PIC: 0.75; ACB: 1.2): high-risk indications are labeled in red to indicate that both PIC and ACB are above the thresholds, and moderate-risk indications are labeled in orange to indicate that both PIC and ACB have one of them exceeding the threshold; the remaining points are marked in gray. The inspiration for adopting this approach to graphics came from \cite{romero2021semen}.}

\label{fig100} 
\end{figure}

\begin{table}[!ht]
\begin{adjustwidth}{-1.5in}{0in} 
\centering
\caption{{\bf Highly malignant association indications.}}
\begin{tabular}{ccc}
\hline
Indicators & ACB & PIC \\ \hline
Envelope invasion: Present & 5.0943 & 1.0000 \\ \hline
Vascular invasion: $\leq$4 vessels & 2.9694 & 0.9802 \\ \hline
Tumor envelope: Incomplete & 2.7288 & 0.9406 \\ \hline
Vascular invasion: $>$4 vessels & 2.5461 & 1.0000 \\ \hline
Margin: Lobulated & 1.8033 & 1.0000 \\ \hline
Trabecular pattern: Present & 1.7747 & 1.0000 \\ \hline
Echogenic foci: Peripheral calcifications & 1.7239 & 1.0000 \\ \hline
Margin: Irregular & 1.6766 & 0.9408 \\ \hline
Echogenic foci: Macrocalcifications & 1.6595 & 0.9500 \\ \hline
Nodule-in-nodule pattern: Present & 1.6539 & 1.0000 \\ \hline
Mean TSH score: [-5.063,-2.102] & 1.6140 & 0.8214 \\ \hline
Mean diagram: $\ge$4 & 1.5835 & 0.9372 \\ \hline
Halo: Uneven thickness halo & 1.5763 & 0.9456 \\ \hline
Max diagram: $\ge$4 & 1.4284 & 0.9119 \\ \hline
BMI: Overweight: & 1.2987 & 0.9618 \\ \hline
Mean TSH score: [-0.551,0.374] & 1.2969 & 0.9945 \\ \hline
Age: [35,45) & 1.2702 & 0.9708 \\ \hline
Echogenic foci: Microcalcifications & 1.2611 & 0.9800 \\ \hline
Echogenicity: Hypoechoic & 1.2535 & 0.9135 \\ \hline
Hashimoto's thyroiditis: Present & 1.2361 & 0.8835 \\ \hline
Mean TSH score: [-2.048,-0.553] & 1.6965 & 0.7897 \\ \hline
Vascularity: Mainly central vascularity & 1.2357 & 1.0000 \\ \hline
TSH tRMSSD: [0.18841,10.1856] & 1.2253 & 0.7870 \\ \hline
\end{tabular}
\label{tab1}
\end{adjustwidth}
\end{table}

In order to visualize the effect of different clinical indications more intuitively, we further plotted the confidence ratios corresponding to the presence or absence of each indication as violin and swarm plots (due to the large number of points, the swarm plots were plotted using random sampling), as is shown in figure 2. In addition the geometric mean of the confidence ratios for each indication across all rule pairs was plotted as a histogram for comparison.

\begin{figure}[!htbp] 
\includegraphics[width=\textwidth]{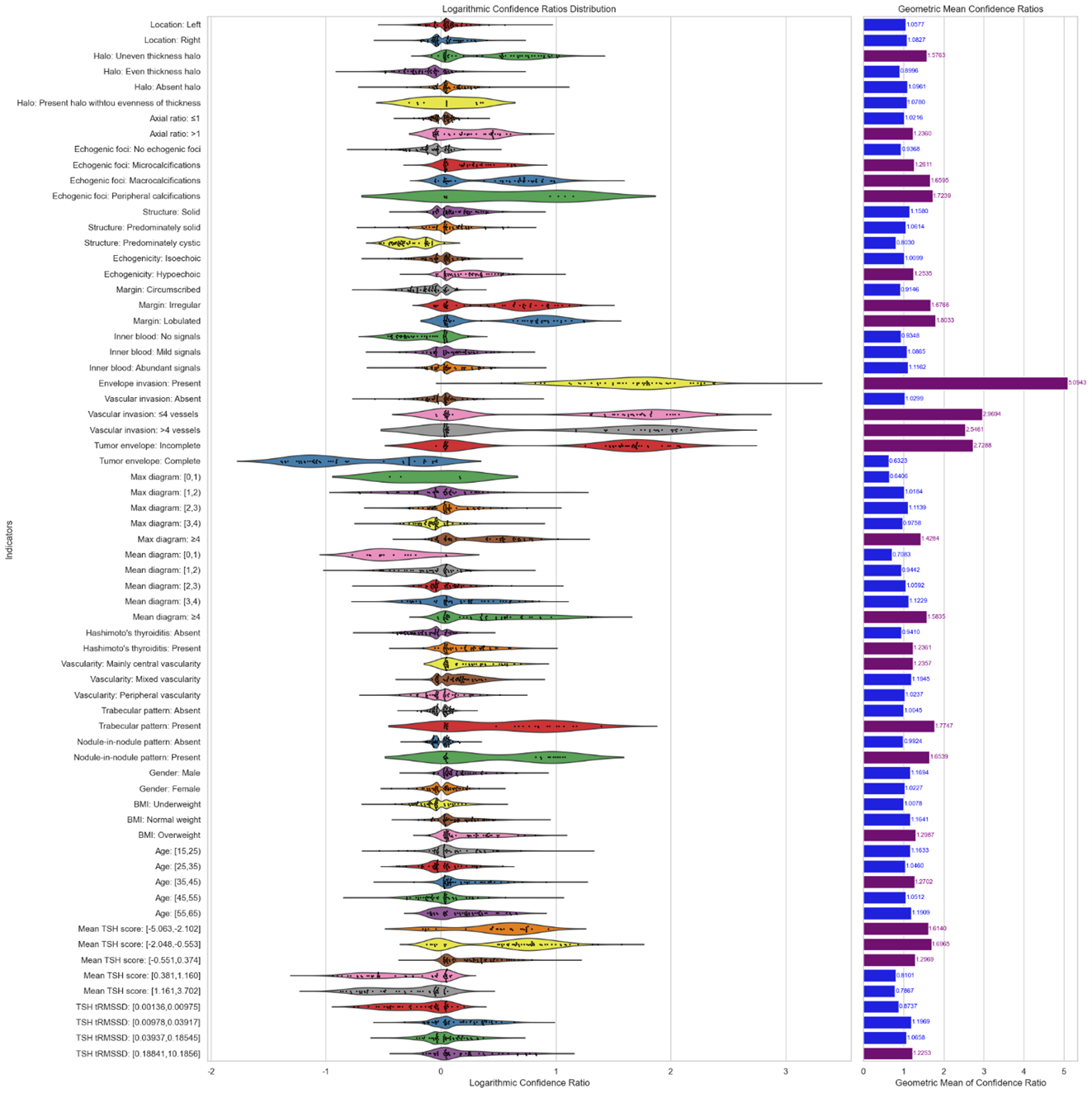}
\caption{\color{Gray} \textbf{Distribution of confidence ratios for different indications} The left panel plots the distribution of confidence ratios corresponding to the presence or absence of different clinical indications. The violin plots are based on all confidence ratios, while the superimposed swarm plots are based on sampling up to 100 points per indication due to the large number of data points. The histogram on the right side shows the geometric mean of all confidence ratios for the different indices, with indices with means above 1.2 labeled in purple and the rest in blue.}
\end{figure}


\section*{Discussion}
Our study observed a significant association between uneven thickness halo and malignancy, which is consistent with previous reports in the literature \cite{zhang2021value}. From the perspective of nodule diagram, observation of Figure 2 shows that the mean diagram presents a clearer trend compared to the maximum diagram, i.e. the larger the mean diagram, the higher the overall confidence ratio, in other words, the greater the risk of malignancy. In addition, both the maximum diagram and the mean diagram exceeding 4 cm are very alarming. Regarding the color Doppler indices, the results in Figure 2 suggest that mainly central vascularity and mixed vascularity have a certain risk of malignancy, which is consistent with previous findings in the literature \cite{zhang2024role}. We also found that infrequently studied signs, such as the trabecular pattern and the nodule-in-nodule pattern, have a very clear association with malignant risk. For BMI, we found a strong malignant association with overweight BMI (as seen in Figure 2, the malignant association became stronger as the BMI index increased in the BMI grading). However, further research is needed here because our dataset covers so few people with obese BMI that the rule containing obese BMI was not picked up by the algorithm's confidence threshold and incorporated into the results. In terms of TSH-related metrics, our results suggest a malignant association between both low mean TSH scores and relatively high TSH tRMSSD.
Our proposed method of calculating ACB has a similar starting point to SHAP, in that it averages the difference in the impact of the presence or absence of the feature of interest on the results in some sense to reflect the impact of the feature. The difference is that our computation comes directly from the direct reflection of the dataset itself rather than the model's prediction, which is more straightforward in terms of interpretability. However, the ACB calculation has its own limitation. In SHAP, all possible combinations of features other than that feature are considered when calculating the SHAP value for a given feature, and then the differences in the model output are weighted and averaged. However, in our study, due to the limitations of association rule mining itself in terms of rule generation and the limited number of feature combinations contained in the dataset itself, we were not able to weight all the possibilities to compute the ACB. We did not weight the confidence ratios according to the probability of occurrence of the feature combinations as SHAP does. However, we set a filtering rule: whether it is the calculation of ACB or the plotting of violin and swarm plots, we filtered out those data points whose confidence ratios were too low. This is based on the consideration that we want to avoid ‘pairs of rules with small changes in confidence’ from smoothing out the average change in confidence when calculating the geometric mean. We therefore set a threshold $\kappa=\log\left(1/0.95\right)\approx0.223$ to exclude from the analysis those data points with confidence ratios below the threshold.
\paragraph{Limitations of the study}
\textbf{Further refinement of the analytical framework.} There is subjectivity in the selection of thresholds when screening for indicators of high malignant associations based on ACB and PIC. This means that for those indications that are in the ambiguous zone, we cannot accurately determine theoretically whether they have significant malignant associations. In the future, we expect to be able to combine this set of analyses with rigorous statistical analyses or machine learning frameworks to give it a stronger theoretical foundation.
\textbf{Algorithm optimization.} Although we optimize the apriori algorithm for association rule mining for specific datasets, the overhead in time when the data is too large is still objective. For example, on the personal laptop we used (CPU: AMD Ryzen 7 7735H 3.20GHz, RAM: 40GB), the two-stage apriori algorithm on the dataset took (150+30)s when we chose $\eta_{supp}=10/1673$, while it took about (900 +150)s. When the dataset to be processed is larger, it may be necessary to choose a more optimized data structure such as FP-tree \cite{han2000mining} to further improve the efficiency of the algorithm.
\textbf{Limitations of the dataset.} Due to the size of the dataset itself, the combination of indicators contained in the data was incomplete, with only 67 of the 100 clinical indications originally recorded (the 2 labels for benign and malignant were excluded) being presented in the results, and a number of indicators, such as ‘BMI: obese’, were excluded from the analysis framework because they did not meet the minimum confidence requirements.
\textbf{Clinical utility.} From the point of view of practical clinical use, the conclusions of this work are still not amenable to direct use. A more representative work in terms of clinical ease of use is the F-TIRADS \cite{li2023us}, and in the future we expect to be able to design more practical preoperative diagnostic aids for follicular thyroid cancer based on the results of the study.


\section*{Acknowledgments}
This work is supported by the National Natural Science Foundation of China (No. 12171275), and Tsinghua University Education Foundation.

\section*{Author Contributions}
All authors contributed to the study conception and design. Material preparation, data collection and analysis were performed by Songlin Zhou, Tao Zhou, Xin Li and Stephen Shing-Toung Yau. The draft of the manuscript was written by Songlin Zhou and Tao Zhou, all authors commented on previous versions of the manuscript.

\nolinenumbers

\bibliography{library}

\bibliographystyle{abbrv}

\end{document}